\documentclass[conference]{IEEEtran}
\IEEEoverridecommandlockouts
\usepackage{cite}
\usepackage{amsmath,amssymb,amsfonts}
\usepackage{hyperref}
\usepackage{float}
\usepackage{algorithm}
\usepackage{algpseudocode}
\usepackage{graphicx}
\usepackage{textcomp}
\usepackage{xcolor}
\def\BibTeX{{\rm B\kern-.05em{\sc i\kern-.025em b}\kern-.08em
    T\kern-.1667em\lower.7ex\hbox{E}\kern-.125emX}}
\begin{document}

\title{A Simple and Effective Method for Uncertainty Quantification and OOD Detection\\

% \thanks{Funding information.}
}

\author{\IEEEauthorblockN{Yaxin Ma}
\IEEEauthorblockA{\textit{Electrical and Computer Engineering} \\
\textit{University of Florida}\\
Gainesville, USA\\
yaxinma@ufl.edu}
\and
\IEEEauthorblockN{Benjamin Colburn}
\IEEEauthorblockA{\textit{Electrical and Computer Engineering} \\
\textit{University of Florida}\\
Gainesville, USA \\
benjamin.colburn@ufl.edu}
\and
\IEEEauthorblockN{Jose C. Principe}
\IEEEauthorblockA{\textit{Electrical and Computer Engineering} \\
\textit{University of Florida}\\
Gainesville, USA \\
principe@cnel.ufl.edu}
}

\maketitle

\begin{abstract}
Bayesian neural networks and deep ensemble methods have been proposed for uncertainty quantification; however, they are computationally intensive and require large storage. By utilizing a single deterministic model, we can solve the above issue. We propose an effective method based on feature space density to quantify uncertainty for distributional shifts and out-of-distribution (OOD) detection. Specifically, we leverage the information potential field derived from kernel density estimation to approximate the feature space density of the training set. By comparing this density with the feature space representation of test samples, we can effectively determine whether a distributional shift has occurred. Experiments were conducted on a 2D synthetic dataset (Two Moons and Three Spirals) as well as an OOD detection task (CIFAR-10 vs. SVHN). The results demonstrate that our method outperforms baseline models.

\end{abstract}


\section{Introduction}
Deep learning has demonstrated success in numerous applications such as computer vision, natural language processing, etc \cite{b1} \cite{b2} \cite{b3}.
However, one significant issue with deep learning is its tendency to produce overconfident predictions. In highly sensitive applications such as autonomous driving or medical-related fields  \cite{b4} \cite{b5}, the consequences of prediction errors can be severe, as these areas are directly linked to human safety and well-being. As a result, uncertainty quantification has emerged as a critical and indispensable area of research to address these challenges effectively\cite{b51}\cite{b52}\cite{b6}.

In general, uncertainty can be categorized into two main types: data uncertainty and model uncertainty \cite{b7} \cite{b8}. Data uncertainty arises from the inherent randomness or noise in the data and cannot be reduced through the training process. Model uncertainty, also known as epistemic uncertainty, is caused by uncertainty in the model's parameters and architecture, reflecting the limitations in the model's knowledge or learning capability \cite{b9}. One type of uncertainty called distributional uncertainty occurs when the distributions of the training set and test set differ. Recently, distributional uncertainty has been tackled by using the model activations to simplify the estimation \cite{b10}.

To quantify uncertainty in the context of distribution shift, there are two primary applications: active learning \cite{b41}\cite{b42}\cite{b43} and out-of-distribution (OOD) detection\cite{b44}\cite{b45}. Active learning selectively chooses the most informative training data and directs data acquisition toward regions of high uncertainty, thereby minimizing the need for excessive labeled data while preserving model accuracy. For OOD detection, the goal is to identify test samples whose distribution differ from the training set. For example, if the training data consists of the MNIST dataset, which contains only digit images, and the test data includes samples from the Fashion MNIST dataset, which contains images of clothing, the model, regardless of how well it is trained, cannot provide correct predictions for such samples. In such cases, OOD detection can identify these test samples as out-of-distribution instances.

Typically, from the perspective of deep learning, the model is trained through optimization, resulting in a fixed set of parameters being selected. In contrast, from a probabilistic viewpoint, Bayesian model averaging considers all possible parameter sets, weighted by their posterior probability\cite{b11}. Bayesian methods utilize Bayes’ theorem to update beliefs about a neural network's parameters based on observed data. However, directly computing the posterior distribution is often infeasible due to the complexity of neural networks. As a result, researchers have developed various techniques to approximate this posterior distribution \cite{b31}\cite{b32}\cite{b33}. Monte Carlo (MC) Dropout provides a practical approach to approximate Bayesian inference by applying different dropout masks during inference and performing multiple stochastic forward passes\cite{b12}. But it has limited uncertainty representation and computational drawbacks.

The ensemble method is another approach for complementing the model output with uncertainty estimation \cite{b13}. It estimates the variance among model outputs as an indicator of uncertainty. However, both Bayesian and ensemble methods demand large computational resources for training. Additionally, ensemble methods require more storage to accommodate the parameters of multiple models. 

To address this issue, researchers have proposed using deterministic methods with a single model for uncertainty quantification\cite{b34}\cite{b35}. By utilizing one model, deep deterministic method try to learn the latent representation of a model or apply a distance-sensitive function to estimate predictive uncertainty \cite{b36} \cite{b37}\cite{b38}. For example, deterministic uncertainty quantification (DUQ) measures uncertainty by computing the distance between the model output and the closest class centroid using an RBF network and gradient penalty\cite{b14}. Spectral-normalized Neural Gaussian Process (SNGP) enhances distance-awareness by applying weight normalization and replacing the output layer with a Gaussian Process\cite{b15}. Deep Deterministic Uncertainty (DDU) approximates the class-conditional distribution using a Gaussian Mixture Model and quantifies uncertainty through log-likelihood estimation\cite{b16}.  

However, these methods rely on specific assumptions which may not always hold in real-world scenarios, potentially limiting their flexibility. DDU assumes that the features of each class follow a Gaussian distribution, which will not accurately represent more complex feature spaces.  In our proposed method, we leverage the Information Potential Field (IPF), a non parametric estimator of the probability density function, to estimate the density of sample projections in feature space without imposing such assumptions. The IPF will quantify the density of the training set which is the in-distribution (iD) data, while the OOD data are the test set samples. Furthermore, unlike DDU, our approach does not require treating each class separately; instead, we compute the feature space density directly across all classes. This results in a simpler and more generalized approach, making it applicable to a wider range of scenarios while maintaining robustness and accuracy.

%Our method offers the following key contributions:

%(a) We applied the concept of the information potential field for uncertainty quantification. The IPF quantifies the density of in-distribution (iD) data. To evaluate a test sample, we place it within the field to assess the amount of in-distribution information it contains. If the test sample is located in a region of high potential, it indicates low uncertainty.

%(b) Our method does not require OOD samples for training the model. We do not pose assumptions for the feature space density and do not rely on class labels when approximating the distribution of features. This enhances the method’s generalizability and applicability to diverse tasks.

We conduct experiments on two 2D synthetic datasets: the two-moons and three-spirals datasets. We are the first to use the three-spirals dataset for visualizing uncertainty performance, as the commonly used two-moons dataset is too simplistic. Additionally, our experimental results on OOD detection using the CIFAR-10 and SVHN datasets highlight the superiority of the proposed method.

\begin{figure*}
    \centering
    \includegraphics[width=1\linewidth]{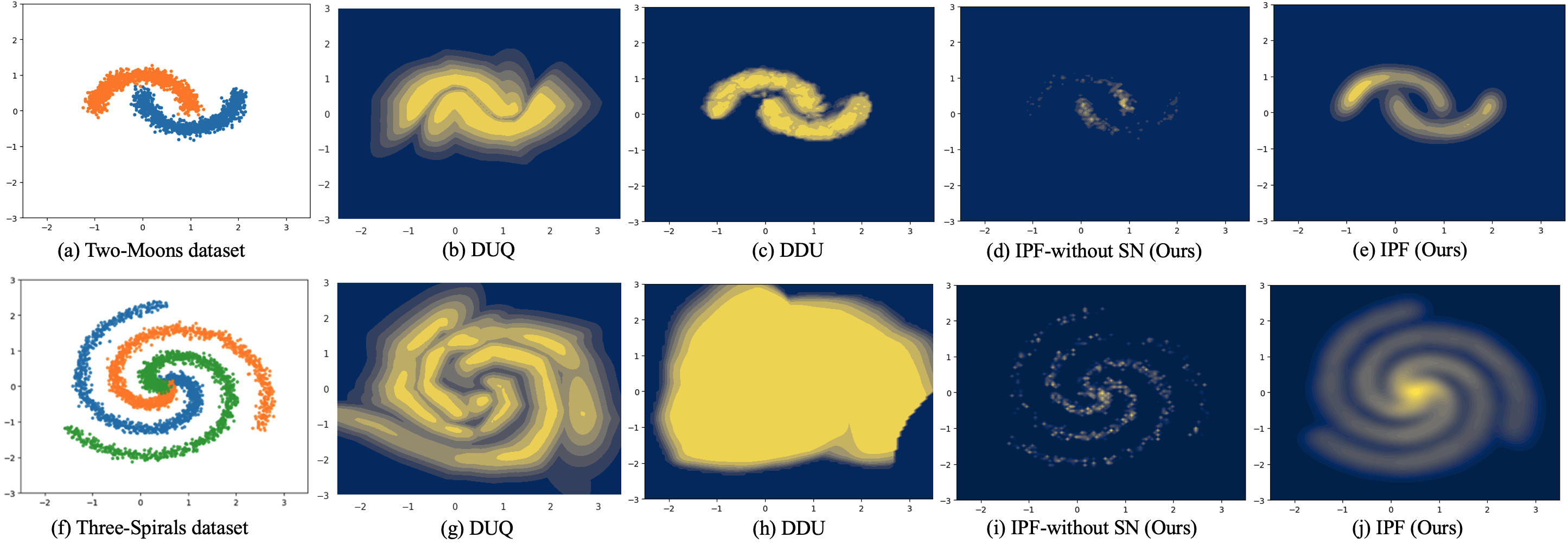}
    \caption{Uncertainty results of different baseline methods on the Two-Moons dataset and Three-Spirals dataset. The first row corresponds to the Two-Moons dataset, and the second row represents the Three-Spirals dataset. DUQ and DDU were selected as baseline methods, and the effect of spectral normalization (SN) was analyzed. The blue region indicates high uncertainty, while the yellow region represents low uncertainty. Ideally, low uncertainty is expected in regions covered by the training data, and high uncertainty in areas outside these regions. For the Two-Moons dataset, the IPF method clearly depicts the central uncertainty region where no training data are present compared to DUQ and DDU. For the Three-Spirals dataset, the IPF method demonstrates a precise and interpretable uncertainty region that aligns closely with the training data.}
    \label{fig:res1}
\end{figure*}

\section{Method}
\subsection{Problem Formulation}

In classification tasks, the underlying assumption is that the distribution of the training set is the same as the distribution of the test set. However, in practical applications, the training set distribution may differ from the test set distribution, resulting in a distributional shift. If a test sample shares the same distribution as the training set, it is referred to as in-distribution (i-D) data. Conversely, if the test sample comes from a distribution different from the training data, it is classified as OOD data. In such cases, the prediction for the test sample carries a high degree of uncertainty, known as distributional uncertainty\cite{b10}\cite{b21}\cite{b22}.

Following this definition, density-based and distance-based methods have been proposed to measure uncertainty\cite{b14}\cite{b16}\cite{b17} \cite{b18}. Specifically, if a test point is located in a high-density region of the training data, or if its distance to the training data is small, the test point is considered to have low uncertainty. Conversely, if the test point lies in a low-density region or is far from the training data, it is deemed to have high uncertainty, as it represents unseen data.

\subsection{Distance Awareness Representation}

For most large datasets, the data space is high dimension and very complex, making it challenging to measure density or distance directly in the raw data space. Neural network models are an efficient approach to extract features. These latent features can be directly used as representations of data, and normally they exist in a lower dimensional space. The activations of the neural network layer before the classification layer, is normally used as the projected data. Therefore, it will be easier to use the feature space density or feature space distance for uncertainty quantification. %Another advantage of using features is that we can evaluate the quality of the trained model for OOD detection. 
%However, this approach mixes distributional and epistemic uncertainty, which complicates the problem.   

In fact, the features learned by the model are not guaranteed to capture sufficient variability or distinguishability within the input data. As a result, it becomes difficult to differentiate between iD and OOD data. For instance, features extracted from OOD data may be mapped into the same regions of the feature space as those learned from iD data. This phenomenon, known as feature collapse\cite{b19}, undermines the effectiveness of using latent features to measure the distance or density of a test sample relative to the training set.

Therefore, we utilize the bi-Lipschitz constant to improve the quality of the the features during model training as suggested in \cite{b15}. For inputs \({x}_1\) and \({x}_2\),
\begin{equation}
L\left\|{x}_1-{x}_2\right\|_\mathcal{X} \leq\left\|f_{\theta}\left({x}_1\right)-f_{\theta}\left({x}_2\right)\right\|_\mathcal{Z} \leq U\left\|{x}_1-{x}_2\right\|_\mathcal{X},
\end{equation}
where \( L \) and \( U \) are constants representing the lower and upper bounds, respectively. The function $f_\theta(\cdot)$ represents the neural network model. \( \left\| \cdot \right\|_{\mathcal{X}} \) and \( \left\| \cdot \right\|_{\mathcal{Z}} \) denote the distances in the sample space and feature space, respectively. 
 
During model training, we employ Spectral Normalization (SN)\cite{b15} to enforce the bi-Lipschitz constraint to improve  the model's ability to maps distinct inputs to distinct representations. 

\subsection{Informational Potential Field}

Let \( D_{\text{train}} \) and \( D_{\text{test}} \) represent the training set and test set, respectively. The training set consists of samples \(\left(x_i, y_i\right)\), where \(x_i \in \mathcal{X}\) and \(y_i \in \mathcal{Y}\), \( \mathcal{X} \) and \( \mathcal{Y} \) denote the sample space and label space. . Similarly, the test set is represented as \(\left(x^*_i, y^*_i\right)\). Let \( f_\theta(x) \) denote a neural network model parameterized by \(\theta\), trained on \(D_{\text{train}}\). We define \( z \in \mathcal{Z} \) as the latent feature representation, where \(\mathcal{Z}\) denotes the latent feature space. For a sample \(x \in \mathcal{X}\), the model maps \(x\) to a latent feature \(z = f_\theta(x)\).

Let \( p(z) \) denote the feature space density of the projected samples at the top layer (before the classification layer) of the trained  model. To estimate the probability density function of the model activations, we employ the information potential field to approximate \( p(z) \), yielding \( \psi(z) \approx p(z) \). The concept of IPF was introduced in \cite{b20} and is inspired by kernel density estimation. It serves as the equivalent of a probability measure in a Reproducing Kernel Hilbert Space (RKHS). Here IPF quantifies the density of in-distribution projected data, forming a field analogous to a gravitational field in physics. Unlike the DDU method, which assumes the feature space density of the training set follows a Gaussian mixture model, our approach does not impose any assumptions on the feature space distribution. Furthermore, our method does not require consideration of individual classes when estimating the feature space density.  

The IPF provides a density field expressed as: 
\begin{equation}
\psi(z) = \frac{1}{N} \sum_{i=1}^N G(z - z_i)    
\end{equation}
where \( G \) is a Gaussian kernel, \( z \) is the point of interest in the test feature space, and \( z_i \) represents the points in the training feature space, irrespective of their class labels. \( N \) denotes the number of samples in the training set.

The field represents the sum of Gaussian functions centered at each training sample, providing an estimation of the probability distribution. When a point in the information potential field has a high value, it indicates greater information, corresponding to a high feature space density. Consequently, the uncertainty at that point is low. In contrast, a low value at a point in the field signifies a low feature space density, providing insufficient information, which results in high uncertainty.

%Given that the representation lies in a high-dimensional space, we utilize a multidimensional Gaussian kernel:
%\begin{equation}
%    k\left(\mathbf{z}, \mathbf{z}_i\right) = \exp \left(-\frac{1}{2}\left(\mathbf{z} - \mathbf{z}_i\right)^T \Sigma^{-1}\left(\mathbf{z} - \mathbf{z}_i\right)\right)
%\end{equation}
%where \(\mathbf{z}\) represents the feature vector of the test sample, and \(\mathbf{z}_i\) denotes the feature vector of a training sample. \(\Sigma\) is the covariance matrix that defines the shape and scale of the Gaussian kernel.

In our case the data is a vector of size given by the dimensionality of the top layer of the neural network. So, we employ the isotropic Gaussian kernel:
\begin{equation}
k\left(\mathbf{z}, \mathbf{z}_i\right) = \exp \left(-\frac{\|\mathbf{z} - \mathbf{z}_i\|^2}{2 h^2}\right)    
\end{equation}
Here, \(h\) is the kernel width, controlling the scale of the Gaussian function. By adjusting \(h\), we can fine-tune the sensitivity of the model for OOD detection. 

We present the algorithm description in Algorithm 1.
\begin{algorithm}[H]
\caption{Information Potential Field}
\textbf{Input:} $(x_i, y_i)$ in $D_{train}, i=1\cdots N, (x^*, y^*)$ in $D_{test}$
\begin{algorithmic}[1]
\State Train the neural network model $f_\theta(x)$ using $D_{train}$
\State Compute the feature representations for the training set: $z=f_\theta(x)$
\State Compute the feature representation for the test sample: $z^*=f_\theta(x^*)$
\State Compute the Information Potential Field (IPF): $\psi(z^*) = \frac{1}{N} \sum_{i=1}^N G(z^* - z_i)$
\If{$\psi(z^*)$ is low (low density)}
    \State \Return Out-of-Distribution
\Else
    \State \Return In-Distribution
\EndIf

% \If{low density $\psi(z^*)$}
%     \State \Return OOD
% \Else{\State \Return in-distribution data} 
\end{algorithmic}
\end{algorithm}

\section{Experiment} 

\begin{figure*}
    \centering
    \includegraphics[width=1\linewidth]{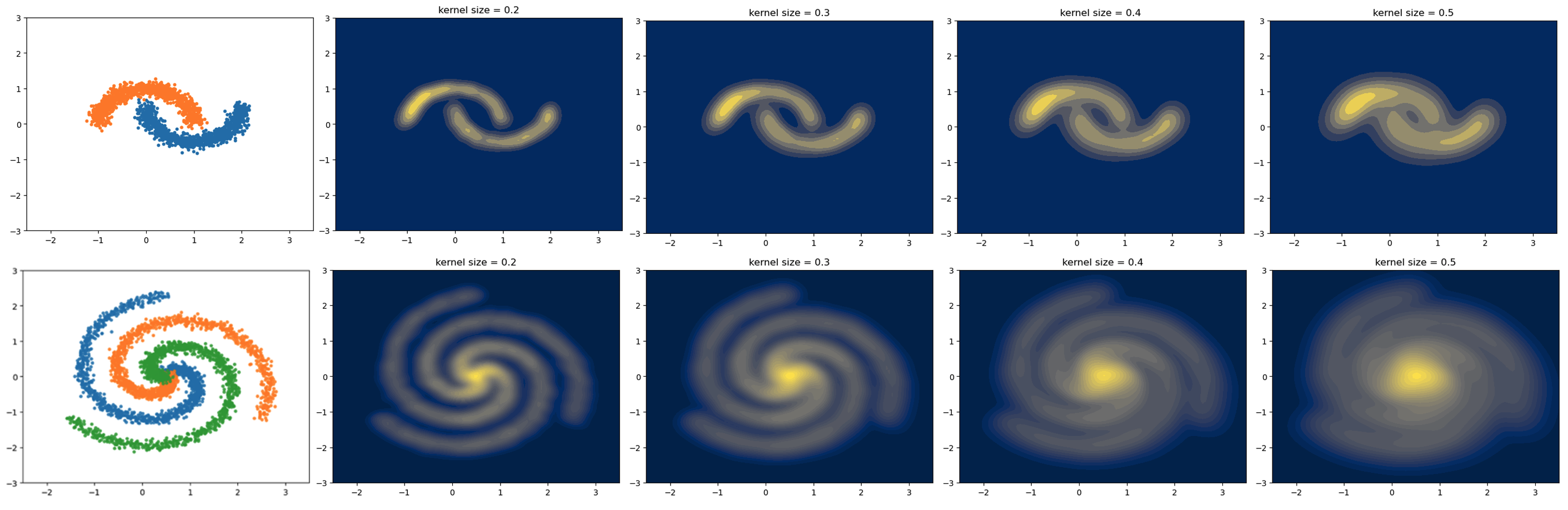}
    \caption{Uncertainty results on the Two-Moons and Three-spirals dataset based on different kernel size. The first row corresponds to the Two-Moons
dataset, and the second row represents the Three-Spirals dataset. Kernel sizes of 0.2, 0.3, 0.4, and 0.5 were applied during the analysis. 
}
    \label{fig:res2}
\end{figure*}

In this section, we first demonstrate the performance of our method on a synthetic 2D dataset and examine the appropriateness of the IPF for OOD detection, and the influence of different kernel sizes. Latter, we also present OOD detection performance and validation with other baseline methods in CIFAR-10 as the in-distribution dataset and SVHN as OOD dataset. 
% The implementation code is available at the following GitHub link: xxxxxxxxxxxxxxx.
%Finally, we perform an ablation study to evaluate the impact of spectral normalization on feature performance.

\subsection{Dataset}
The Two-Moon Dataset is a synthetic dataset consisting of two interlocking half-circle shapes (moons), where each moon represents a different class. For visualization purposes, we set the x-axis range from -2.5 to 3.5 with 100 values and the y-axis range from -3 to 3 with 100 values.

The Three-Spiral Dataset is another synthetic dataset. It contains three intertwined spirals, each representing a different class. For visualization, we use the same range and number of values as the Two-Moon Dataset.

The CIFAR-10 Dataset is a widely used benchmark for image classification algorithms. It consists of 60,000 color images, each sized 32×32, categorized into 10 classes, with 6,000 images per class. The classes include: Airplane, Automobile, Bird, Cat, Deer, Dog, Frog, Horse, Ship, and Truck\cite{b61}.

The SVHN (Street View House Numbers) Dataset is a real-world image dataset designed for digit recognition tasks. It comprises over 600,000 32×32 RGB images of house numbers, extracted from Google Street View imagery\cite{b62}. 

\subsection{2D synthetic data - two moons, three spirals}

Our method was initially evaluated using the two moons benchmark dataset, where 2,000 samples per class were generated with a Gaussian noise standard deviation of 0.1. To further demonstrate the effectiveness of the IPF method, a new three-class spiral dataset was employed, consisting of 1,200 samples for each class with a Gaussian noise level of 0.08.

We compared our method against DUQ\cite{b14} and DDU\cite{b16}, both of which use feature space density or feature space distance to quantify uncertainty. Following \cite{b15}\cite{b16}, the model architecture consisted of a four-hidden-layer neural network with residual connections, where each layer had 128 neurons. We used the top 128 layer activations in our study. The model was trained for 300 epochs using SGD as the optimizer. We use the same model for both the two-moons and three-spirals dataset.

The hyper parameter in the IPF method is the kernel size, which has to be determined from data either using the Silverman's rule of thumb \cite{b73} or using cross validation. In our tests we employed a range of kernels between 0.1 to 1 and the best kernel size we select is 0.3.

Fig. \ref{fig:res1} shows the distributional uncertainty for both the two moons dataset and the three spirals dataset. Ideally, we expect low uncertainty in regions covered by training data and high uncertainty outside these regions. For the two moons dataset, a noticeable gap exists in the central region where no training data is located. Compared to DUQ and DDU, our method distinctly highlights this area with a clear and distinguishable blue region, indicating high uncertainty due to the absence of training data. Compared with DUQ, our method presents a cleaner, more concise uncertainty region without covering areas where no data exists. This results in a more precise and interpretable shape.

For the three spirals dataset, the DUQ method provides a rough estimation of uncertainty, while DDU struggles with this dataset due to its Gaussian prior assumption for each class. In contrast, our method produces better results, accurately aligning uncertainty regions with the shape of the training data, demonstrating superior performance.

The impact of spectral normalization on the proposed method is also evaluated. The second-to-last column in Fig. \ref{fig:res1} represents our method without SN, while the last column shows our method with SN. It is evident that incorporating spectral normalization leads to significant improvements, demonstrating that the proposed method, which integrates IPF with spectral normalization, is highly effective for uncertainty quantification.

We also conducted experiments to assess the impact of different Gaussian kernel sizes on uncertainty quantification. Fig. \ref{fig:res2} presents the results obtained with kernel sizes of 0.2, 0.3, 0.4, and 0.5. Adjusting the kernel size allows effective control over the model's sensitivity and tolerance to uncertainty. This highlights the flexibility and adaptability of our method.

\begin{table*}[htbp]
\caption{OOD detection results of different baselines for Wide-ResNet-28-10 ON CIFAR-10}
\begin{center}
\renewcommand{\arraystretch}{1.3}
\begin{tabular}{|c|c|c|c|}
\hline
\textbf{Method} & \textbf{Accuracy $(\uparrow)$} & \textbf{ECE} $(\downarrow)$ & \textbf{AUROC (SVHN as OOD)}$(\uparrow)$ \\
\hline
Softmax & 93.18 $\pm$ 0.009 & 0.031 $\pm$ 0.010 & 85.65 $\pm$ 0.013  \\
\hline
Ensemble\cite{b13} & 94.80 $\pm$ 0.001 & 0.007 $\pm$ 0.001 & 91.95 $\pm$ 0.015 \\
\hline
DUQ\cite{b14} & 93.85 $\pm$ 0.002 & 0.048 $\pm$ 0.010  & 92.43 $\pm$ 0.005\\
\hline
DDU\cite{b16} & 93.15 $\pm$ 0.009 & 0.030 $\pm$ 0.010 & 92.90 $\pm$ 0.016 \\
\hline
Our method (IPF)& 93.38 $\pm$ 0.008  & 0.028 $\pm$ 0.008 & 93.18 $\pm$ 0.006 \\
\hline
\end{tabular}
\label{tab1}
\end{center}
\end{table*}

\begin{figure}
    \centering
    \includegraphics[width=0.7\linewidth]{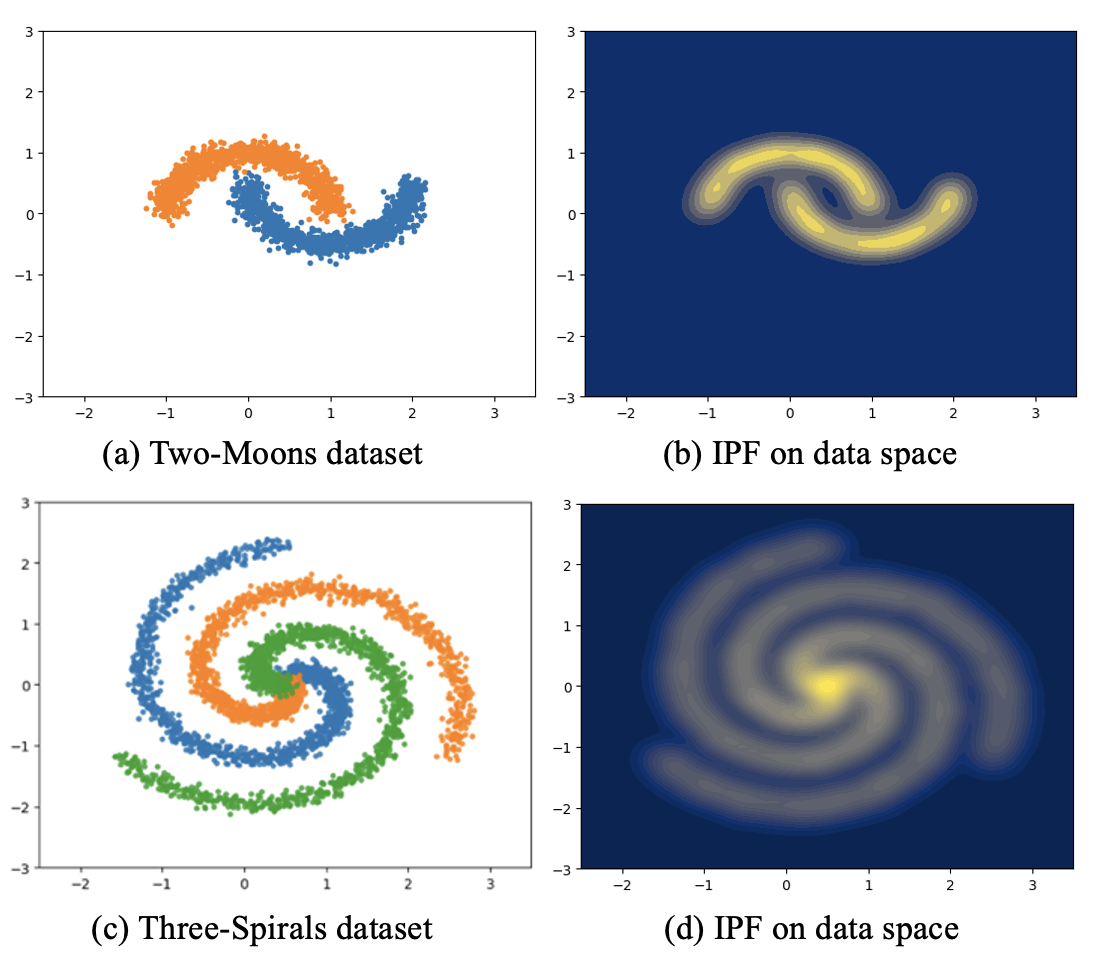}
    \caption{Uncertainty results for the Two-Moons dataset and Three-Spirals dataset using IPF directly applied in the data space.}
    \label{fig:res_data}
\end{figure}

\subsection{OOD detection}

We employ the CIFAR-10 dataset as the in-distribution (iD) data to train the model, and its performance is evaluated using classification accuracy and the Expected Calibration Error (ECE)\cite{b63}. The SVHN dataset is chosen as OOD data due to its distribution significantly differs from that of CIFAR-10. During the inference phase, both iD and OOD data are selected as the test set. After extracting the test set features from the trained model, we apply the proposed IPF method to quantify uncertainty. For each test sample, if it is located in a high-density region of the feature space corresponding to the iD data, it indicates low uncertainty and is classified as iD data. Conversely, if the test sample has a very low feature space density value relative to the training data, it indicates high uncertainty and is classified as OOD data.

To evaluate the model’s OOD detection performance, we treat OOD detection as a binary classification task, where iD data represents one class and OOD data represents the other. The Area Under the Receiver Operating Characteristic curve (AUROC) is used as the evaluation metric.

The proposed method was compared against several popular baselines for uncertainty quantification, such as softmax, ensemble, DUQ, and DDU. The softmax entropy of a standard deep neural network was chosen as a simple baseline. For the ensemble method, we configured it to consist of five models with the same architecture but trained using different parameters. 

For our training model, we employed the Wide ResNet-28-10 architecture \cite{b64}\cite{b65}, where the numbers 28 and 10 denote the model's depth and width, respectively. We utilized the layer preceding the final fully connected layer as the feature embedding, with a feature dimension of 640. To determine the optimal kernel width, we performed cross-validation over the range [0.01, 1], selecting the value that maximized the AUROC score. The best kernel width selected was 0.35. To ensure consistent and reliable results, each experiment was conducted five times. The outcomes were then averaged, and the standard deviation was computed to assess variability. Our experiments were conducted using the PyTorch framework and executed on four NVIDIA A6000 GPUs. The training was conducted for 350 epochs, with SGD as the optimizer. The learning rate was initialized at 0.01 with a decay schedule, and the batch size was set to 1,024.

The comparison results of different baselines with IPF method are presented in Table 1. Note that in our evaluation, AUROC serves as the metric for assessing OOD detection performance. As shown in the table, our method achieves a higher AUROC score, clearly demonstrating superior performance of the proposed IPF approach. Additionally, the IPF method is much simpler to apply since it does not require the presentation of each class individually. We were surprised with the quality of the Parzen estimator for the size of the layers, which means that more advanced IPF estimators will improve the results even further.  

\subsection{IPF in the Input Space}

%In the experiment conducted on the 2D synthetic dataset, we evaluated the performance of our method both with and without spectral normalization. The results of this experiment are illustrated in Fig. 1. The second-to-last column represents our method without SN, while the last column depicts our method with SN. A significant difference is obvious between the results obtained with and without SN, demonstrating that the application of spectral normalization enhances the feature extraction capability for uncertainty quantification.
We also evaluate the proposed IPF method applied directly to the data space. Specifically, instead of using a neural network to obtain embeddings, we approximate the density of the training set directly from the raw data and evaluate the test samples within this density. As shown in Fig. \ref{fig:res_data}, it is evident that applying IPF in the data space achieves results comparable to those in the feature space, provided the dimensionality of the data space is small. Additionally, we conducted experiments using the IPF method on the CIFAR-10/SVHN dataset, but the performance significantly dropped compared to IPF in the feature space, which was expected because of the data dimension. Note that in this case the dimensionality of the data space is huge (32x32) so the IPF computed with radially symmetric Gaussians, which corresponds to the Parzen window method is not appropriate.

% We selected the best kernel size for the high dimension space that turned out to be XXXXX. The AUROC score obtained was xxx, which is lower than that achieved in the feature space.

%We also employed t-SNE\cite{b66} to visualize the feature space of in-distribution data and OOD data. Fig. \ref{fig:tSNE} presents the experimental results, where the orange points correspond to one class (iD data) and the blue points represent another class (OOD data). The visualization reveals that indeed portions of the iD data points can be distinguished from those of OOD data, but that there is still a lot of overlap between the representations of the two data sets.

%\begin{figure}
   % \centering
    %\includegraphics[width=0.6\linewidth]{img/res3.png}
    %\caption{Feature space visualization using t-SNE. The orange points represent in-distribution data from CIFAR-10, while the blue points correspond to OOD data from SVHN.}
   % \label{fig:tSNE}
%\end{figure}

\section{Discussion and Conclusion}
From the above experiment, the proposed IPF method can be directly applied to the data space if the dimensionality is not too high. This can speed up tremendously the OOD detection. However, for image datasets like CIFAR-10, we are unable to achieve results comparable to those obtained in the feature space. Neural network models provide an efficient way to extract high-level features, reducing the dimensionality and making these features easier to process further. We show that the IPF can improve state of the art results for OOD detection.  However, using features extracted from neural network models introduces a mixture of distributional and epistemic uncertainty. This is because the process of model training and embedding extraction inherently involves epistemic uncertainty related to the model parameters, which complicates the problem. Future studies on quantifying distributional uncertainty should focus more on data-centric approaches and aim to minimize the impact of uncertainty introduced by model training.

The difficulties of the IPF estimation were expected because for probability density function estimation in high-dimensional data, Parzen estimation does not scale well, and in practice it should not be used above 20 dimensions\cite{b20}. In our current work, we use the isotropic Gaussian kernel estimator due to its simplicity and as part of our preliminary investigation. There are more advanced methods that extend probability density estimation up to 500 dimensions\cite{b71}\cite{b72}.  Incorporating these advanced methods willl be pursued in future research because they could, on one hand, enable more accurate density approximation in the feature space and, on the other hand, potentially allow IPF to be applied directly in the data space on image datasets. This would eliminate the epistemic uncertainty introduced during model training and simplify the challenges of distributional uncertainty and OOD detection.

In summary, we have developed an effective approach for quantifying uncertainty in distributional shifts for OOD detection. By leveraging the information potential field, we achieved a more realistic approximation of the feature space density. Our experiments are conducted on 2D synthetic datasets, including the two-moons and three-spirals datasets, as well as OOD detection tasks comparing the CIFAR-10 and SVHN datasets. The experimental results highlight the superiority of the proposed method while the IPF methodology was the simplest (Parzen estimation). Future work will demonstrate the utility of advanced RKHS methods in this line of research.  

\section*{Acknowledgment}

This research was supported by the grant N000142312571 and N000142512223.

\end{document}